\documentclass{INTERSPEECH2023}

\interspeechcameraready 
\usepackage{multirow}
\usepackage{makecell}
\usepackage{stfloats}
\usepackage{subfigure}
\usepackage{cite}
\title{ Improving Code-Switching and Named Entity Recognition in ASR with Speech Editing based Data Augmentation}
\name{Zheng Liang, Zheshu Song, Ziyang Ma, Chenpeng Du, Kai Yu, Xie Chen$^\dagger$ \thanks{ $\dagger$ Corresponding author}}
\address{
  MoE Key Lab of Artificial Intelligence, AI Institute \\ 
X-LANCE Lab, Department of Computer Science and Engineering \\ 
Shanghai Jiao Tong University, Shanghai, China}
\email{\{liangzhenglz, songzheshu, zym.22, duchenpeng, kai.yu, chenxie95\}@sjtu.edu.cn}

\begin{document}

\maketitle
 
\begin{abstract}
Recently, end-to-end (E2E) automatic speech recognition (ASR) models have made great strides and exhibit excellent performance in general speech recognition. However, there remain several challenging scenarios that E2E models are not competent in, such as code-switching and named entity recognition (NER). Data augmentation is a common and effective practice for these two scenarios. However, the current data augmentation methods mainly rely on audio splicing and text-to-speech (TTS) models, which might result in discontinuous, unrealistic, and less diversified speech. To mitigate these potential issues, we propose a novel data augmentation method by applying the text-based speech editing model. The augmented speech from speech editing systems is more coherent and diversified, also more akin to real speech. The experimental results on code-switching and NER tasks show that our proposed method can significantly outperform the audio splicing and neural TTS based data augmentation systems.
\end{abstract}
\noindent\textbf{Index Terms}: speech recognition, speech synthesis, data augmentation, code-switching, named entity recognition 
\section{Introduction}
\label{sec:intro}
In recent years, end-to-end (E2E) automatic speech recognition
(ASR) models have gained increasing research interest and present superior performance compared to conventional hybrid systems\cite{ref_on_address,le2021contextualized,deng2021alleviating,winata2020adapt}. Nevertheless, there remain several challenging scenarios for E2E ASR models, which might limit their potential applications, such as code-switching and named entity recognition tasks.

Code-switching is a common language phenomenon where two or more languages appear in one utterance. The recognition of code-switching is important but challenging for E2E ASR models. There are always insufficient code-switching audios in the training data, while E2E ASR models are data-driven and their performance heavily depends on the diversity and coverage of the content of training data. 
Named entity recognition faces a similar dilemma for E2E ASR models. It is impractical to collect abundant training data to cover all possible names of persons, organizations, places, and other named entities. Consequently, the recognition performance of named entities and code-switching is far worse than that of common words in standard E2E ASR systems. 
Both tasks suffer from the problem of data scarcity, as it is impractical to collect sufficient training data to cover all possible cases of code-switching and named entities. 
Moreover, both tasks involve rare words and long tail \cite{le2021contextualized,deng2021alleviating,winata2020adapt} distributions where high-frequency words contribute to most of the appearance, while low-frequency words are rarely observed, which are hard to model by E2E ASR systems.

There are consistent and active research efforts on E2E models aiming to improve code-switching and named entity recognition performance given the importance of these two application scenarios. 
For the code-switching, there is a series of previous works in the E2E ASR model from the perspective of specific model design and explicit language information incorporation~\cite{ref_cs_ctc_1,ref_cs_ctc,ref_cs_ctc_2,ref_cs_ctc_att,ref_cs_att,ref_joint_cs,ref_cs_streaming}. 
On the other hand, for named entity recognition, various aspects have been explored \cite{ref_ner_beamsearch,ref_ner_shallow_fusion,ref_clm,ref_clas,ref_ner_fast,ref_contextual} in order to boost named entity recognition. 
One simple approach is to apply the phrase biasing technique by adding additional rewards on the named entity list during decoding \cite{ref_ner_beamsearch,ref_ner_shallow_fusion}.  
Another recent research direction is to treat the given named entity list as contextual information that can be integrated into the E2E ASR model during training and decoding \cite{ref_clas,ref_ner_fast,ref_contextual}. 
Data augmentation provides another popular and effective solution for both code-switching and named entity recognition in E2E ASR models by generating paired audio and text data, which is also the main research topic in this paper. There are mainly two data augmentation approaches in literature \cite{ref_on_address,ref_cpd,ref_cs_speech_oov,ref_cs_aug_using_tts,ref_using_syn_audio,ref_adaptation_rnnt,ref_improving_rnnt, ref_cpd_aug}. 
One approach is to apply advanced TTS models to generate audio given the text of interest. One potential drawback of using the TTS model is the artifact and lacking diversity in acoustic and speaker characteristics from the synthesized audio. The other approach is to splice audio segments with small units, such as phone and word, from the existing corpus to form a long audio corresponding to the given text. However, this concatenative operation might introduce discontinuity and inconsistency.
Therefore, there still exists clear discrepancies between the audios generated from current data augmentation approaches and real recordings.

In order to mitigate the discrepancy between augmented and real audio, in this paper, we propose a novel data augmentation method by applying the text-based speech editing technique. 
By ``editing'' the real audio, the augmented speech with the proposed method is able to retain the consistency and diversity of recording audio, and is expected to match  well with the unseen audio in test time.
The experimental results demonstrate that our proposed data augmentation yields significant performance improvements over competitive baseline systems, including neural TTS and audio splicing based data augmentation methods, on both code-switching and named entity recognition tasks.
\vspace{-0.2cm}
\section{Related work}
\label{sec:related}
\subsection{Audio splicing and neural TTS for data augmentation}
\label{ssec:splice_tts}
In this section, we briefly review two common types of speech data augmentation approaches, which are audio splicing and neural TTS.

Audio splicing is a straightforward and effective method to form new audio by simply concatenating speech segments of small units, such as phonemes and words, from the training corpus. One advantage of this method is that it preserves most of the characteristics of real speech. Audio splicing has been widely adopted for data augmentation to enhance speech recognition tasks, especially for code-switching and named entity recognition scenarios~\cite{ref_on_address,ref_cpd,ref_cs_speech_oov}. 
Moreover, for the target words that are scarce in the training data, the more fine-grained spliced segments or neural TTS can be applied to generate the speech segments of such words. 
However, one practical issue of audio splicing methods is the inconsistency of the segments from different sources in the augmented audio, which might sacrifice the potential performance gain.

With the recent advance of neural TTS models \cite{ref_fastspeech2, ref_gradtts, ref_cpd_vqtts}, high-fidelity audio can be easily generated. Therefore, neural TTS can be naturally applied for data augmentation to synthesize audios of given text \cite{ref_cpd,ref_cs_speech_oov,ref_cs_aug_using_tts,ref_using_syn_audio}. 
Unlike audio splicing which relies on existing speech data, neural TTS models generate speech with high quality and naturalness by converting text input into speech audio.
However, most neural TTS models aim to produce high-quality audio, instead of audio with complicated and diversified acoustic and speaker conditions. The artifact and lacking diversity in synthesized audio might limit the performance improvement for ASR tasks. Hence, in order to achieve better performance, most neural TTS based data augmentation methods need to mix synthesized data with similar amounts of real data  to alleviate the impact of acoustic mismatch \cite{ref_cs_aug_using_tts,ref_using_syn_audio,ref_adaptation_rnnt}.
In this paper, we use the non-autoregressive TTS model, FastSpeech 2~\cite{ref_fastspeech2}, to generate high-quality augmented audios in a fast inference speed, and HiFiGAN~\cite{ref_hifigan} is applied as the vocoder for waveform generation based on the predicted Mel-spectrogram from FastSpeech 2.
\vspace{-0.2cm}
\subsection{Text-based based speech editing}
\label{ssec:speech_edite}
Text-based speech editing is a technique that can ``edit'' audio by deleting, inserting or replacing the specified words in a flexible way \cite{ref_editspeech,ref_a3t,ref_campnet,ref_speech_painter,ref_zero_shot_tts,ref_retriever_tts}. The ``edited'' audio retains similar speaker, channel and prosody attributes as the original audio. Hence, the speech editing model has the potential to generate audio akin to the recording audio in the training corpus, while easing the aforementioned issues from audio splicing and neural TTS methods. In the next section, we will elaborate on the use of speech editing for data augmentation.
\vspace{-0.2cm}
\section{Data augmentation with speech editing}
\label{sec:aug}
To overcome the above-mentioned drawbacks of existing data augmentation methods and generate speech with better authenticity and consistency, we propose a novel data augmentation by applying the text-based speech editing model on the original audios from the training corpus, to generate augmented audio with specified modification. 

\begin{figure}[htbp]
\centering
\centerline{\includegraphics[scale=0.5]{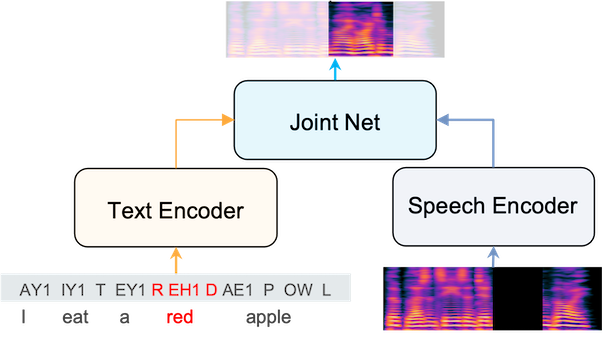}}
\vspace{-0.2cm}
\caption{Overall framework of the text-based speech editing model applied in this paper. The model is trained to generate the speech according to the edited text and the corresponding original audio. }
\vspace{-0.5cm}
\label{fig:overview_model}
\end{figure}

We extend the FastSpeech 2 with the BERT-style contextual modeling \cite{ref_bert} that empowers the ``extraneous'' speech segment to capture the rich contextual information of the recorded audio, including speaker and prosody features, for text-based speech editing~\footnote{The code and samples of the speech editing model are available at https://liangzheng-zl.github.io/bedit-web}.
The speech editing model architecture is illustrated in Figure \ref{fig:overview_model}. In the model, 
a text encoder is applied to capture the content information, a speech encoder is constructed to capture the acoustic attributes, and these two pieces of information are fused in the joint net to yield the Mel-spectrogram for further speech generation. 
By incorporating acoustic information derived from authentic speech, the speech editing model can generate speech that exhibits a high degree of contextual coherence and diversity in the modified region. 
\vspace{-0.2cm}
\vspace{-0.2cm}
\subsection{Augmentation of code-switching recognition}
\label{ssec:aug_cs}
\begin{figure}[htbp]
	\centering
    \subfigure[Speech editing for code-switching scenario]{
		\begin{minipage}[b]{\linewidth}
            \centering
			\includegraphics[width=\linewidth]{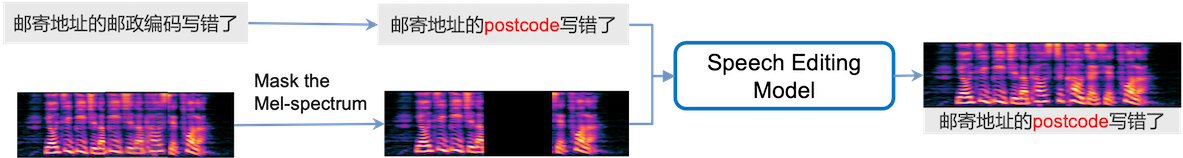}
		\end{minipage}
		\label{fig:se_cs}
	}\\
 \vspace{-0.2cm}
	\subfigure[Speech editing for named entity scenario]{
		\begin{minipage}[b]{\linewidth}
            \centering
			\includegraphics[width=\linewidth]{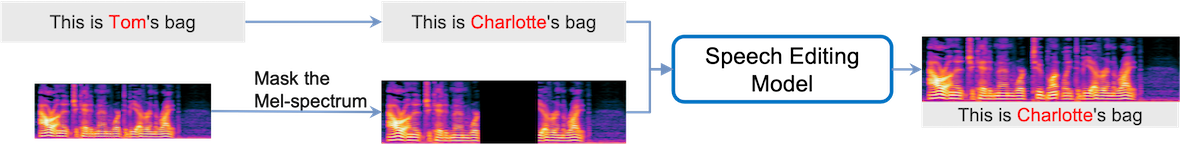}
		\end{minipage}
		\label{fig:se_ner}
	}
    \vspace{-0.3cm}
	\caption{The pipeline of data augmentation by applying the text-based speech editing model for code-switching and named entity scenarios.}
	\label{fig:speech_editing}
 \vspace{-0.5cm}
\end{figure}

The steps of generating code-switching speech with the proposed method, taking Mandarin-English for example, are as follows: 
First, forced alignment is performed on the existing monolingual Mandarin data to obtain the phone-level alignment information between the text and speech, as well as the boundaries of the words.
Then, the English word or phrase is inserted or replaced into the Mandarin text following the grammatical rules of both languages to generate the code-switching text. Moreover, there are a series of previous works \cite{ref_cs_text_aug_1,ref_cs_text_aug_2,ref_cs_text_aug_3,ref_cs_text_aug_4} aiming to generate code-switching text data. 
And then, the speech editing model predicts the duration of the modified region, i.e. the English words. The mask is then applied to the specified region of the Mel-spectrogram with some fixed value, e.g. 0.1. The well-trained model generates the Mel-spectrogram of the masked region based on the text information and contextual acoustic information captured from the edited text and masked Mel-spectrogram, and concatenates the generation with the original Mel-spectrograms to obtain edited Mel-spectrograms. 
Finally, the edited Mel-spectrogram is directly outputted or inputted into a vocoder to synthesize speech. We illustrate the pipeline of the model in Figure \ref{fig:se_cs}.
\vspace{-0.2cm}
\subsection{Augmentation of named entity recognition}
\label{ssec:aug_ner}
The generation of the speech with named entities is similar to the generation of code-switching speech. Specifically, given the named entities list and the texts with modified positions, the edited text can be easily obtained by insertion and replacement in the specified position. And masks are applied on the recorded audio, where the length of masking can be estimated from the model. Inference can be performed on the speech editing model given the complete text and masked speech, to obtain the edited Mel-spectrogram. The general process is illustrated in Figure \ref{fig:se_ner}.

\vspace{-0.2cm}
\section{Experiments}
\label{sec:exp}
We conduct experiments on two distinct tasks: code-switching and named entity recognition in ASR. 
For both tasks, all the audio is sampled to 16kHz and extracted 80-dimensional log-Mel filterbanks (Fbank) with 50ms frame size, 12.5 ms frame hop, and the Hann window function. The force alignment and grapheme-to-phoneme are both carried out by the GMM-HMM model implemented on Kaldi toolkit~\cite{ref_kaldi} to convert words to phones and acquire alignment information between the audio segments and phones, and the neural TTS and E2E ASR models are built on the ESPnet toolkit~\cite{ref_espnet}.
\vspace{-0.2cm}
\subsection{Code-switching in ASR}
\label{ssec:cs_exp}
\subsubsection{Datasets}
\label{sssec:cs_exp_data}
The code-switching speech recognition experiments are conducted on the ASRU 2019 Mandarin-English Code-Switching data \cite{ref_asru},  which contains 200 hours of code-switching speech by 566 speakers and 500 hours of Mandarin speech by 1880 speakers. Additionally, 460 hours of mono-English data by 1172 speakers from the LibriSpeech dataset~\cite{ref_librispeech} are also included. The test set for the experiments comprises  20 hours of code-switching speech.
To compare the augmented speech with the real speech, we randomly sample 100h data from the 200h code-switching data as $Aug_{real}$ set to serve as the simulation of the real collected code-switching data for data augmentation.
 The other half of 100h code-switching data combined with two kinds of monolingual data are used as $Train$ set.
\vspace{-0.2cm}
\subsubsection{Models and tasks}
\label{ssec:cs_tasks}
We apply the multilingual ASR model LAE~\cite{ref_lae} without any language model for experiments. 
 To compare the effectiveness of the proposed method with other methods, besides the $Aug_{real}$ set, five other augmented sets are prepared with the same texts in the $Aug_{real}$ set for data augmentation as follows.
1)$Aug_{splice}$ set: The speech is generated with the audio splicing method. The English word speech segments in the sentences of the $Aug_{real}$ set are cut according to the alignment information. Then, FastSpeech 2 and HiFiGAN synthesize the English word speech segments with the inputs of the words. Finally, the synthesized speech segments of the words are inserted into the corresponding positions in the original speech to generate the speech data.
2) $Aug_{tts}$ set: The speech data is generated by the FastSpeech 2 and HiFiGAN with the text data from the $Aug_{real}$ set.
3) $Aug_{edit}$ set: The speech data is generated by the text-based speech editing model and HiFiGAN. The regions of the English words in the original speech from the $Aug_{real}$ set are masked and then fed into the speech editing model with the corresponding texts to generate the speech.
4) $Aug_{edit-feats}$ set: The process of generating this set is similar to that of $Aug_{edit}$. However, we keep only output Fbank feats generated by the speech editing model instead of converting them to waveform with HiFiGAN.
5) $Aug_{edit-feats+tts}$ set: Half of the augmented data is from $Aug_{edit-feats}$, the other half is from $Aug_{tts}$.

The ASR baseline system is trained on the $Train$ set. Different combinations of data augmentation sets ($Aug_{real}$, $Aug_{splice}$, $Aug_{tts}$, $Aug_{edit}$, $Aug_{edit-feats}$, $Aug_{edit-feats+tts}$) and $Train$ are used to train ASR model separately. We evaluate these models using mixed error rate (MER), word error rate (WER) for English and character error rate (CER) for Mandarin.
To keep the consistency of the speaker, FastSpeech 2, HiFiGAN and the speech editing model are all trained on the 100h code-switching speech in $Train$ set.
\subsubsection{Results}
\label{ssec:cs_results}
The results of the code-switching test set are summarized in Table \ref{tab:cs_results}. It can be observed that all the data augmentation methods can help to improve the performance of the ASR model on all the metrics, and the best scores on all metrics are achieved by training with $Train$+$Aug_{real}$. In particular, training with $Train$+$Aug_{edit}$, $Train$+$Aug_{edit-feats}$ and $Train$+$Aug_{edit-feats+tts}$ yields similar results to training with $Train$+$Aug_{real}$, and the CER and MER of training with $Train$+$Aug_{edit-feats}$ achieve the same scores with $Train$+$Aug_{real}$. This indicates that our proposed method can generate speech with consistency and diversity analogous to real speech.
Moreover, our proposed method can bring a higher boost in data augmentation results than the other two methods. 
The impact of the vocoder is shown by comparing the results of training with $Train$+$Aug_{edit}$ and $Train$+$Aug_{edit-feats}$. It can be seen that the performance in terms of all metrics is consistently degraded by the artifacts introduced by the vocoder. Furthermore, by comparing the results of training with $Train$+$Aug_{edit-feats}$ and $Train$+$Aug_{edit-feats+tts}$, it is revealed that a large proportion of unrealistic synthetic speech may adversely affect the model's performance.
\vspace{-0.15cm}
\begin{table}[htb]
\centering
\captionof{table}{MER, Man(CER) and Eng(WER) results on Code-Switching test set. The model is trained with $Train$ set and augmented sets by different methods (real collection, audio splicing, neural TTS, and text-based speech editing).}\label{tab:my_label}
\resizebox{\linewidth}{!}{
\begin{tabular}{cl||c|c|c} \hline
\multicolumn{2}{l||}{Training data} &  Man CER($\%$) $\downarrow$ & {Eng WER($\%$) $\downarrow$} & {MER($\%$) $\downarrow$} \\ \hline\hline
\multicolumn{2}{l||}{$Train$}  & 9.6 & {33.2} & {11.8}\\ \hline
 & $+Aug_{real}$   & \textbf{8.6}  & \textbf{30.0} & \textbf{10.5}\\
 & $+Aug_{splice}$   & 9.2  & 32.2 & 11.1\\
 & $+Aug_{tts}$   & 9.3  & 31.2 & 11.1\\
 & $+Aug_{edit}$    & 8.8 & 30.7 & 10.7\\
 & $+Aug_{edit-feats}$  & \textbf{8.6} & 30.2 & \textbf{10.5}\\
 & $+Aug_{edit-feats+tts}$  & 8.7 & 30.4 & 10.6 \\ \hline
\end{tabular}
\label{tab:cs_results}
}
\end{table}
\vspace{-0.15cm}
\begin{table*}[!t]
    \centering
    \caption{Overall WER, Recall and Precision results on Giga-name-test (up) and NER-name-test (down). The model is fine-tuned with augmented data by different methods (audio splicing, neural TTS, text-based speech editing), and mixed data with real data (without names) sampled from the training set of ASR model.}
    \vspace{-0.2cm}
     \resizebox{0.79\linewidth}{!}{
    \begin{tabular}{c||c|c|c||c|c|c}
    \hline
    \multirow{3}*{\makecell[c]{Fine-tune \\ data}} & \multicolumn{3}{c||}{\textbf{Aug-data only}} &
    \multicolumn{3}{c}{\textbf{Mixed-data }}\\
    \cline{2-7}
     &  \multirow{2}*{\makecell[c]{Overall\\ WER($\%$) $\downarrow$ }}  & \multicolumn{2}{c||}{person name} & \multirow{2}*{\makecell[c]{Overall\\ WER($\%$) $\downarrow$ }}  & \multicolumn{2}{c}{person name} \\
     \cline{3-4} \cline{6-7}
     &  & Recall($\%$) $\uparrow$ & Precision($\%$) $\uparrow$ & & Recall($\%$) $\uparrow$ & Precision($\%$) $\uparrow$ \\
     \hline\hline
    \multicolumn{7}{c}{Test set: Giga-name-test} \\
    \hline
     \multicolumn{1}{l||}{(w/o fine-tuning)} & 17.6 & 51.0 & 52.0 & - & - &  \\
     \hline
     \multicolumn{1}{c||}{real-data (w/o names)} & - & - & - & 17.5 & 51.2 & 52.1 \\
     \hline
     \multicolumn{1}{l||}{$Aug_{splice}$} & 15.9 & 73.7 & 74.2 & 15.9 & 73.1 & 73.5 \\
     \multicolumn{1}{l||}{$Aug_{tts}$} & 16.0 & {75.4} & {75.3} & 15.7 & 74.9 & 75.0\\
     \hline
     \multicolumn{1}{l||}{$Aug_{edit}$} & 15.8 & 74.4 & 74.7 &  15.6 &  75.5 & 75.7 \\
     \multicolumn{1}{l||}{$Aug_{edit-feats}$} & 15.7 & 74.7 & 75.0 &  15.6 &  74.2 & 74.7 \\
     \multicolumn{1}{l||}{$Aug_{edit-feats+tts}$} & \textbf{15.5} & \textbf{78.7} & \textbf{78.9} &  \textbf{15.5} &  \textbf{78.1} & \textbf{78.2} \\
     \hline\hline
      \multicolumn{7}{c}{Test set: NER-name-test} \\
    \hline
      \multicolumn{1}{l||}{(w/o fine-tuning)} & 12.0 & 53.0 & 52.6 & - & - & - \\
     \hline
     \multicolumn{1}{l||}{real-data (w/o names)} & - & - & - & 11.7 & 53.0 & 52.4 \\
     \hline
     \multicolumn{1}{l||}{$Aug_{splice}$} & \textbf{11.2} & {63.3} & {63.0} & {11.1} & {63.5} & {62.9}\\
     \multicolumn{1}{l||}{$Aug_{tts}$} & 12.3 & 65.3 & 64.4 & 11.3 & 66.3 & 65.3 \\
     \hline
    \multicolumn{1}{l||}{$Aug_{edit}$} & 11.7 & 65.0 & 64.3 & 11.2 & 65.8 & 65.0 \\
    \multicolumn{1}{l||}{$Aug_{edit-feats}$} & \textbf{11.2} & 65.5 & 65.0 & 11.0 & 65.5 & 64.6 \\
    \multicolumn{1}{l||}{$Aug_{edit-feats+tts}$} & \textbf{11.2} & \textbf{73.8} & \textbf{72.7} & \textbf{10.9} & \textbf{72.7} & \textbf{71.7} \\
    \hline
    \end{tabular}
    }
\vspace{-0.4cm}
    \label{tab:ner_in_domain}
\end{table*}
\vspace{-0.4cm}
\subsection{Named entity recognition in ASR}
\label{ssec:ner_exp}
\subsubsection{Datasets}
\label{ssec:ner_exp_data}
In the task of named entity recognition in ASR, we use personal names as the target named entities.
The ASR model is trained on the GigaSpeech-M set, which consists of 1,000 hours of speech audio from different sources including audiobooks, podcasts and YouTube \cite{ref_giga}. And two test sets are prepared as follows.
The first test set, referred to as the Giga-name-test set, is generated by processing the standard GigaSpeech dev and test sets through an open-source NER toolkit~\footnote{https://huggingface.co/dslim/bert-base-NER-uncased}. Specifically, we randomly sample 1,888 sentences that contain person names, resulting in a total of 1,434 unique names and a percentage of 8.3\% of names among all words. This test set serves as an in-domain evaluation of the proposed method's performance in recognizing person names.
The second test set, referred to as the NER-name-test, is extracted from a publicly available dataset released in \cite{ref_ner_test}, which consists of 150 hours of speech audio with 38,891 unique named entity tokens of person, location, and organization. 
For our evaluation purposes, we only use the subset of the test set that contains person names, which comprises a total of 1,650 unique names and a percentage of 5.0\% of names among all words and consists of about 5 hours of audio. It serves as an out-of-domain evaluation to test the generalization ability of our proposed method.

The FastSpeech 2, HiFiGAN and speech editing model are all trained on the HiFiTTS dataset \cite{ref_hifitts}, which consists of 292 hours of speech audio by 10 speakers.
\vspace{-0.15cm}
\subsubsection{Models and tasks}
\label{ssec:ner_model_training}
In this section, we follow the recipe for GigaSpeech in the ESPnet to build the ASR model with 12 conformer blocks of the encoder and 6 transformer blocks of the decoder. 

The target name list is formed based on all names appearing in the test sets. To generate the texts containing the target names, 800 sentences containing person names are randomly sampled from the Gigaspeech-XL set to build the database of the templates. The sentence templates with name tags and names from the name list can be flexibly combined to form complete sentences, which are then used as input for the three methods of speech generation.
Similar to the experiments of code-switching,  but only five augmented sets ($Aug_{splice}$, $Aug_{tts}$, $Aug_{edit}$, $Aug_{edit-feats}$, $Aug_{edit-feats+tts}$) are prepared with the database of the templates and the name list. Finally, the generated speech is used to fine-tune the baseline system. Additionally, about 4 hours of synthesized speech data are used for in-domain testing and about 5 hours for out-domain testing. 
In order to balance the ASR performance between the general and target domains, the synthesized speech is mixed with an equal amount of real speech data, referred to as real-data set and containing no names, randomly sampled from the training set.
During the fine-tuning, we follow the instruction of \cite{ref_on_address,ref_adaptation_rnnt}, where the lower 11 layers of the encoder network are frozen.
We evaluate these models using general WER on all texts, and the recall and precision of the names.
\vspace{-0.15cm}
\subsubsection{Results}
\label{sssec:ner_results}
The results on the Giga-name-test and NER-name-test sets are summarized in Table~\ref{tab:ner_in_domain}. The results of fine-tuning with real-data set show that the data sampled from the training set has little impact on the name recognition performance of the model, which changes marginally compared with the results of the model without fine-tuning.
 It can be observed that fine-tuning with $Aug_{edit-feats+tts}$ achieves the best performance on both in-domain and out-domain test sets. In particular, the results of fine-tuning with $Aug_{edit}$ and $Aug_{edit-feats}$ achieve lower WER and similar or better recall and precision of names than the $Aug_{splice}$ and the $Aug_{tts}$, which show the effectiveness of the proposed method for data augmentation in named entity recognition in ASR. 
 For the results of the Giga-name-test set, with the setting of fine-tuning with mixed data, $Aug_{edit}$ outperforms the $Aug_{edit-feats}$ and the $Aug_{tts}$ performs better than the $Aug_{splice}$. A possible explanation can be attributed to the data diversity and complementary effect introduced by the synthesized speech data from the neural TTS or vocoder. 
 However, using the synthesized speech alone leads to overfitting of the ASR model. By fine-tuning with mixed data, the ASR model balances acoustic characteristics between synthesized and real speech. Similarly, in the ${Aug_{edit-feats}}$ set, the real acoustic features of the unmodified regions of the sentence balance the synthesized acoustic features of the modified regions, and together with the coherence and continuity of the generated acoustic features, ${Aug_{edit-feats}}$ achieves a better balance between the results of WER and the results of name recognition than $Aug_{splice}$, $Aug_{tts}$ and $Aug_{edit}$. Moreover, with the combination of more diversity introduced by the synthesized speech in $Aug_{tts}$, the $Aug_{edit-feats+tts}$ achieves the best score on all metrics.
 The WER and the recall and precision of name trends are consistent in the other NER-name-test set, and the data augmentation with speech editing significantly outperforms baseline systems as shown in Table~\ref{tab:ner_in_domain}.
\vspace{-0.2cm}
\section{Conclusions}
\vspace{-0.15cm}
The recognition of code-switching and named entity is challenging but important for E2E ASR models. Data augmentation provides a feasible and effective approach for these two scenarios. However, there exist several mismatches between the real audio and the augmented speech from the existing data augmentation methods.
In this paper, we proposed a novel data augmentation method by applying the text-based speech editing technique, to directly modify the content of original speech from the training corpus. The augmented audio contains better diversity and is more realistic and coherent than that of traditional data augmentation, such as audio splicing and the neural TTS methods. Significant performance improvements can be achieved on both code-switching and named entity recognition tasks by applying the proposed data augmentation method.
\vspace{-0.2cm}
\section{Acknowledgements}
\vspace{-0.15cm}
This work was supported by the National Natural Science Foundation of China  (No. 62206171), and Alibaba Group through Alibaba Innovative Research Program.

\bibliographystyle{IEEEtran}
\bibliography{sections/mybib.bib}

\end{document}